\documentclass[10pt]{article}
\usepackage[utf8]{inputenc} % This might not be needed with XeLaTeX or LuaLaTeX
\usepackage[arabic,english]{babel} % For Arabic language support

\usepackage[a4paper,textwidth=18cm,textheight=24cm,top=2.85cm, bottom=2.85cm, left=1.5cm, right=1.5cm]{geometry}

\usepackage{template/icdp2009}

\makeatletter
\long\def\@makecaption#1#2{%
\vskip\abovecaptionskip
\sbox\@tempboxa{#1. #2}%
\ifdim \wd\@tempboxa >\hsize
#1. #2\par
\else
\global \@minipagefalse
\hb@xt@\hsize{\box\@tempboxa\hfil}%
\fi
\vskip\belowcaptionskip}
\makeatother

\usepackage{lmodern}

\usepackage{graphicx}
\usepackage{times}

\begin{document}
\noindent

\bibliographystyle{plain}

\title{Hunayn: Elevating Translation Beyond the Literal}

\author{\normalsize Nasser Almousa, King Saud University, Saudi Arabia\\ \normalsize Nasser Alzamil, King Saud University, Saudi Arabia\\ \normalsize Abdullah Alshehri, Prince Sattam University, Saudi Arabia\\ \normalsize Ahmad Sait, King Abdulaziz University, Saudi Arabia}

\maketitle 

%\keywords
%Maximum 5 keywords placed before the abstract.

\abstract
This project introduces an advanced English-to-Arabic translator surpassing conventional tools. Leveraging the Helsinki transformer (MarianMT), our approach involves fine-tuning on a self-scraped, purely literary Arabic dataset. Evaluations against Google Translate show consistent outperformance in qualitative assessments. Notably, it excels in cultural sensitivity and context accuracy. This research underscores the Helsinki transformer's superiority for English-to-Arabic translation using a Fus’ha dataset.

\section{Introduction}
In today's interconnected world, language barriers continue to hinder effective communication across diverse cultures and regions. Machine translation systems like Google Translate have played a significant role in breaking down these barriers. Yet, a closer look reveals that while Google Translate can offer a basic understanding between English and Arabic, its translations often miss the nuanced intricacies of the rich literary heritage of the Arabic language. The complexity of Arabic's grammar, syntax, and cultural context presents a formidable challenge for machine translation to overcome. This gap between machine-generated translations and the depth of literary Arabic underscores the need for a translation tool that goes beyond surface-level translations and captures the true essence of the language.

\section{Related Work}
In recent times, the development of neural network-based translation models, exemplified by the success of transformer architectures (Vaswani et al., 2017), has brought significant advancements to the field. These models show promise in grasping the finer shades of meaning and have contributed to enhancing the quality of translations. Additionally, customising these models with our tailored dataset has proven effective in addressing translation challenges specific to language pairs.

\begin{figure}
    \centering
    \includegraphics[width=8cm, height=8cm]{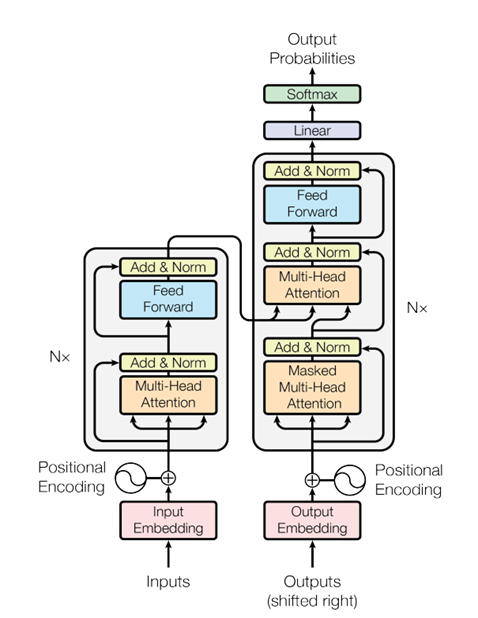} 
    \caption{The Transformer - model architecture}
    \label{fig:plot}
\end{figure}

\section{Methodology}
We present a novel approach to English-to-Arabic translation, leveraging the Helsinki transformer (MarianMT) from Hugging Face. Our methodology includes the collection of a diverse and carefully curated dataset that better reflects the intricacies of the Arabic language to train the model to only output true literary Arabic. We fine-tuned Helsinki on self-scraped data from “Rasaif.com.” Through rigorous evaluation, we demonstrate that our approach outperforms Google Translate, indicating a significant stride toward achieving more accurate and culturally sensitive translations.

\begin{figure}
    \centering
    \includegraphics[width=8.5cm, height=2cm]{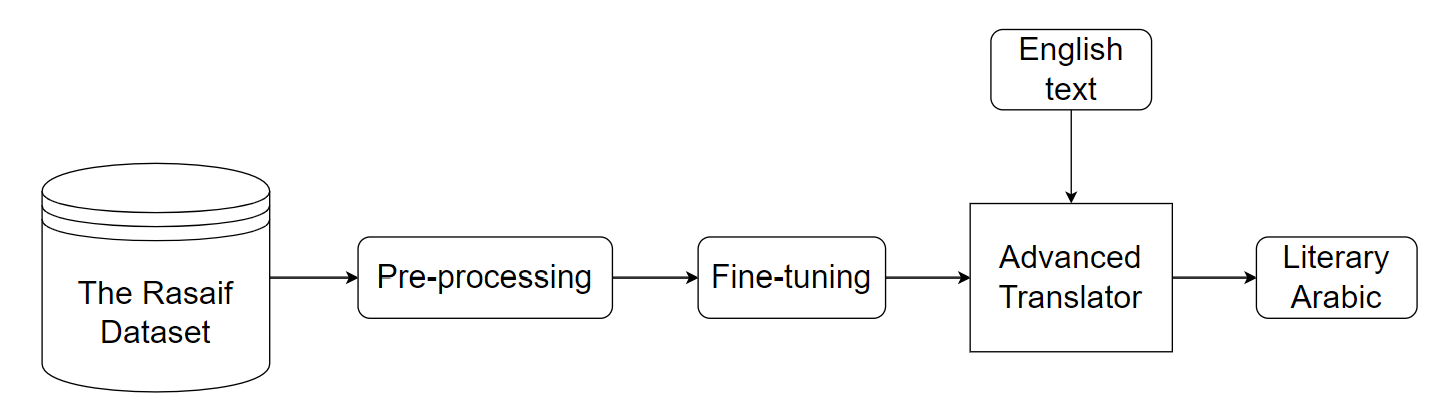} 
    \caption{Methodology}
    \label{fig:plot}
\end{figure}

\section{Dataset}
The dataset used for this project was sourced from the Rasaif website, a repository of old Arabic books that have been translated into English. This selection was made deliberately due to the inherent linguistic and stylistic characteristics of these books, all of which are written in Fus’ha (Classical Arabic). This deliberate choice ensures that the resulting translator outputs exclusively in Fus’ha, aligning with the project's objectives. 

\begin{figure}
    \centering
    \includegraphics[width=8.5cm, height=5cm]{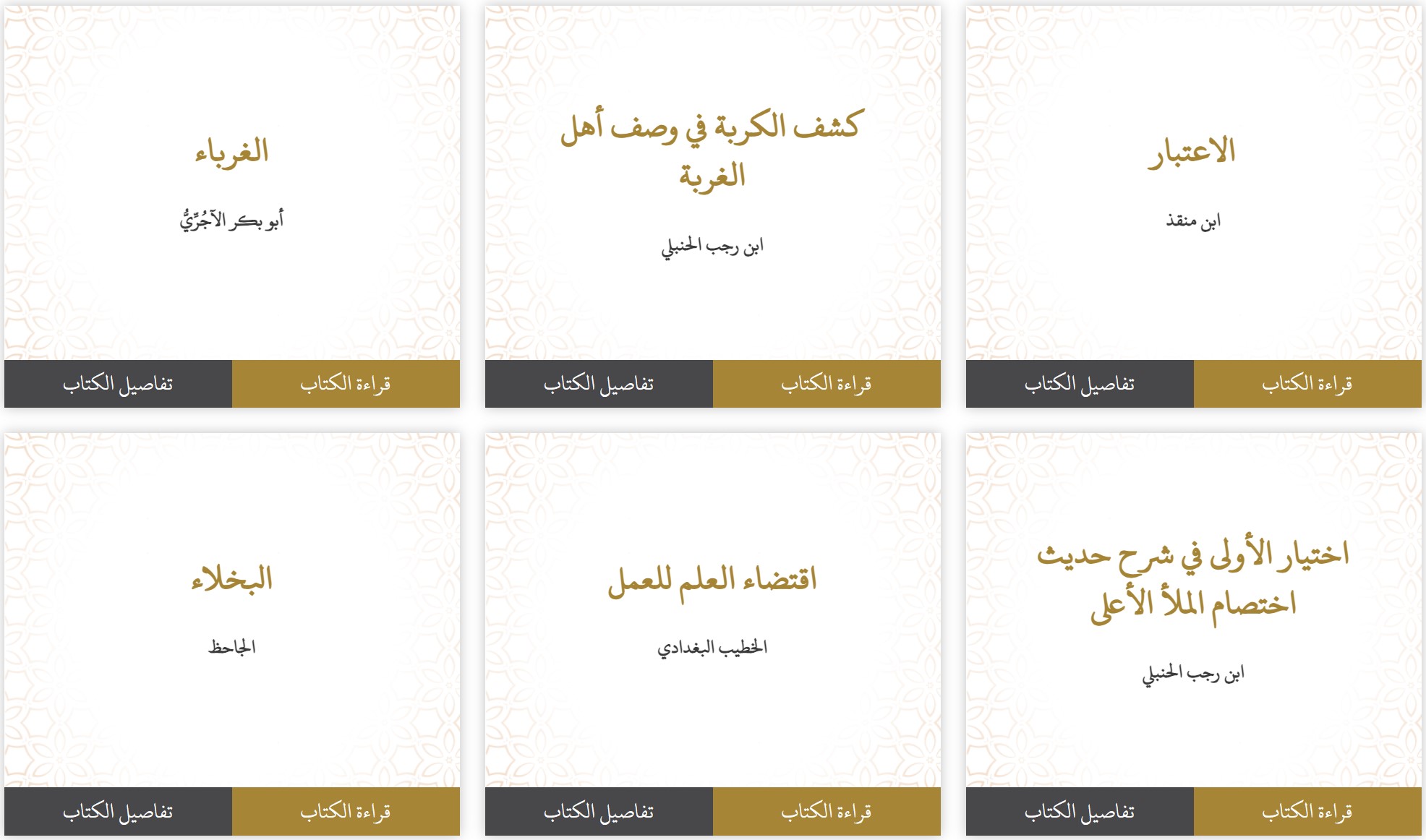} 
    \caption{Examples of books that were scraped.}
    \label{fig:plot}
\end{figure}

\subsection{Data Collection and Preprocessing}
Data preprocessing played a pivotal role in the project, allowing for the creation of a robust and high-quality training dataset. The Rasaif website provided a valuable collection of texts; however, considerable effort was invested in scraping, cleaning and preparing the data to ensure its suitability for training the Helsinki transformer.

\begin{figure}
    \centering
    \includegraphics[width=8.5cm, height=2cm]{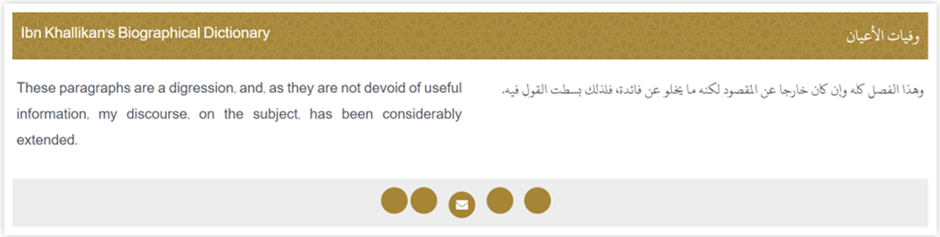} 
    \caption{An Arabic paragraph and its corresponding translation in English.}
    \label{fig:plot}
\end{figure}

\subsection{Cleaning Process and Challenges}
The cleaning process proved to be a demanding task, necessitating meticulous attention to detail. One of the major challenges was the presence of Arabic phrases within the texts that lacked direct English translations and the presence of Arabic diacritics in some of the books. These issues emerged as a critical factor affecting the model's performance, as untranslated phrases disrupted the alignment between the English and Arabic segments of the dataset. And the presence of Arabic diacritics creates inconsistency in the data.
\subsection{Handling Missing Data and Rows}
To address this challenge, significant time was dedicated to data cleaning to eliminate any instances of untranslated Arabic phrases. Data rows that contained such phrases were identified and subsequently removed from the dataset. This strategy aimed to enhance the cohesiveness of the dataset and to mitigate potential issues arising from misaligned bilingual content.
\subsection{Addressing Irregular Data Entries}
In some cases, data entries were found to deviate from the expected content structure. Instead of containing meaningful text, certain rows consisted of an Arabic phrase that meant that the translated text is not in the version of the book and it lacked linguistic significance. Recognizing the importance of maintaining a high-quality dataset, these irregular entries were meticulously removed.

\subsection{Dedication to Data Quality}
The substantial investment of time and effort in data cleaning underscores the project's commitment to producing a high-quality training corpus. The removal of untranslated phrases, missing data, Arabic diacritics, and irregular entries contributed to the creation of a more coherent and linguistically accurate dataset.

While the data cleaning process significantly improved the quality of the training dataset, it is acknowledged that the presence of untranslated phrases and the necessity to remove certain rows might have affected the model's overall performance. Despite these challenges, the resulting dataset serves as a crucial foundation for training the Helsinki transformer and achieving the project's goals.

By addressing these challenges head-on and implementing a rigorous data preprocessing approach, the project aimed to maximise the potential of the Helsinki transformer to deliver accurate and culturally sensitive translations in Fus’ha.

\section{Experimental Setup}
\subsection{Hardware and Software Specifications}
The experimental setup played a crucial role in the successful implementation and evaluation of the advanced English-to-Arabic translator. The computational resources used for training and evaluation were integral to achieving the desired results.
\subsubsection{GPU Selection}
The training process was conducted on a Tesla T4 GPU, a high-performance graphics processing unit optimised for deep learning workloads. This choice of GPU ensured efficient training times and the ability to handle the computational demands of the Helsinki transformer architecture.
\subsubsection{Software Environment}
The experiment was conducted in a virtual environment that utilised the PyTorch and MarianMT libraries. PyTorch provided the foundational framework for implementing machine learning models, while MarianMT offered the tools required to work with the Helsinki transformer model and perform fine-tuning.
\subsection{Comparison Baseline: Google Translate}
To evaluate the performance of the advanced English-to-Arabic translator, a comparison was made against Google Translate, a widely-used machine translation tool. Google Translate served as a baseline to gauge the improvements achieved by the proposed model.
\subsection{Evaluation Metrics}
One metric was employed to comprehensively assess the performance of the advanced translator. This primary qualitative metric was conducted to evaluate the model's cultural sensitivity and context accuracy along with expert evaluation.
\subsection{Training Process}
The training process of Hunayn involved the fine-tuning of the Helsinki transformer architecture using the prepared dataset. Hyperparameters, including batch size, learning rate, and number of training iterations, were optimised through iterative experimentation to achieve the best possible translation performance.
\subsection{Computational Considerations}
Due to the computational demands of training a transformer-based model, careful consideration was given to memory usage and training times. Efficient data loading, memory management, and model checkpointing strategies were employed to maximise training throughput and stability.

\section{Results}

\subsection{Qualitative Assessment: Cultural Sensitivity and Context Accuracy}
We conducted a qualitative assessment of the translations generated by our advanced translator in comparison to Google Translate. This assessment focused on two critical aspects: cultural sensitivity and context accuracy. The translations produced by our model demonstrated a remarkable level of cultural sensitivity. Phrases, idiomatic expressions, and contextually sensitive content were accurately translated, contributing to a more authentic and culturally respectful communication.
\subsubsection{Contextual Accuracy}
Hunayn exhibited a higher degree of context accuracy in comparison to Google Translate. The model's translations effectively captured the nuances and subtleties of the source text, resulting in translations that felt more coherent and contextually appropriate.
\subsection{Arabic Expert Evaluation}
The translated output of the advanced translator was presented to an Arabic language expert, a professor with extensive linguistic knowledge. The expert's input provided a valuable real-world evaluation of the translations' quality, cultural accuracy, and linguistic nuances.
\subsubsection{Expert Feedback}
The expert's feedback was recorded, highlighting specific strengths and areas of improvement observed in the translated output. The expert's insights lent credibility to the project's claims about the translator's efficacy in handling complex linguistic and cultural aspects.
\subsection{Examples: Translations Showcase}
To provide a concrete understanding of the differences in translation quality, we present a selection of example sentences that were translated by ChatGPT, Google Translate, and Hunayn. These examples highlight the strengths of our model in terms of linguistic accuracy, cultural nuances, and context preservation.

\begin{figure}
    \centering
    \includegraphics[width=8.7cm, height=10cm]{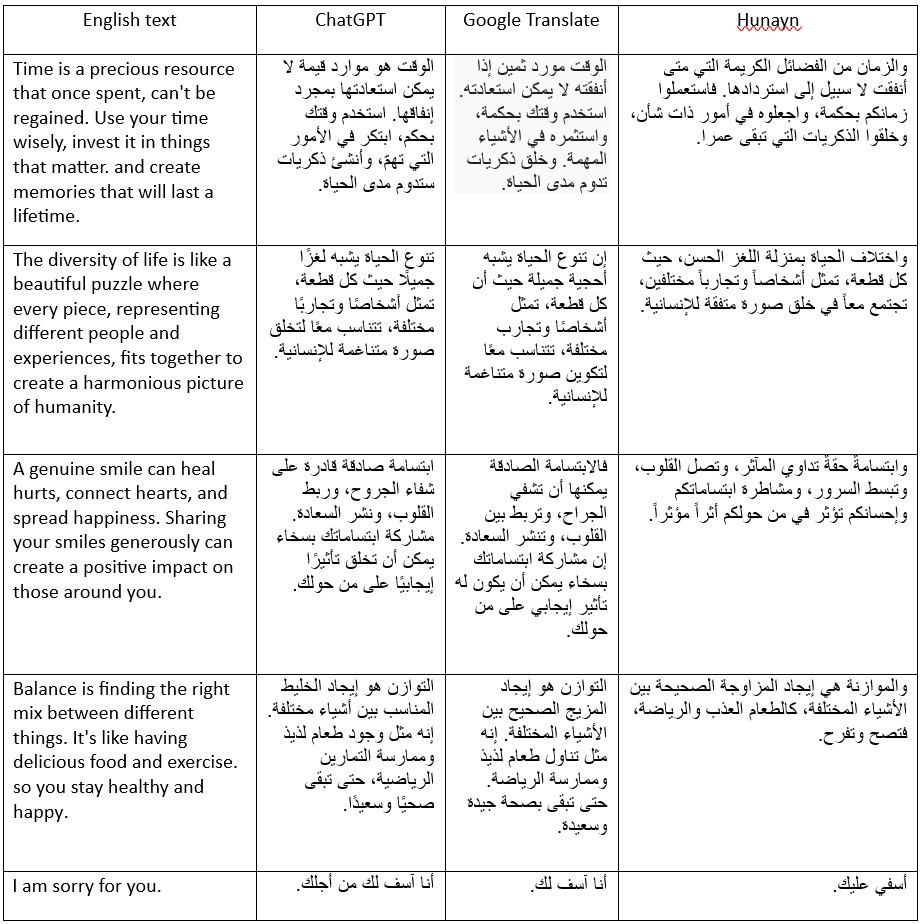} 
    \caption{Comparing translations of conventional tools against Hunayn's.}
    \label{fig:plot}
\end{figure}

\section{Discussion}
\subsection{Model Performance and Improvement}
The results obtained highlight the advanced translator's superior performance compared to Google Translate. The qualitative assessments emphasise the translator's success in maintaining cultural sensitivity and context accuracy, indicating its potential for improving cross-cultural communication.

These results are generated by careful hyperparameter tuning. To output a translation for short sentences like "I am sorry for you" requires the decrease of the "length\_penalty" hyperparameter. We also discovered that increasing the "num\_beams" hyperparameter resulted in more reliable output but inference time takes longer.

\subsection{Challenges and Dataset Impact}
While the advanced translator demonstrated promising results, challenges in the dataset, including untranslated phrases and irregular data entries, impacted its overall quality. The extensive data cleaning process mitigated some of these challenges, but certain instances may have influenced the model's performance. Future improvements could involve refining the data collection and preprocessing strategies to further enhance the dataset's quality.
\subsection{Expert Evaluation and Real-World Applicability}
The involvement of an Arabic language expert in the evaluation process added a crucial real-world dimension to the assessment. The expert's feedback affirmed the model's linguistic accuracy and cultural sensitivity, reinforcing its practical utility for translating literary Arabic texts. This expert validation enhances the credibility of the translator's capabilities in real-world scenarios.
\subsection{Implications and Future Directions}
The success of this project holds implications for cross-cultural communication and translation technology. The advanced translator's ability to produce accurate translations with cultural nuance has the potential to foster deeper understanding between English and Arabic speakers. Future directions could involve expanding the model to other dialects of Arabic, addressing dialectal variations, and incorporating user feedback for continuous improvement.
\subsection{Contributions to Translation Technology} 
This research underscores the significance of leveraging advanced transformer models, such as the Helsinki transformer, for improving English-to-Arabic translation quality. The project's methodology, combining fine-tuning on a carefully prepared dataset with expert evaluation, sets a precedent for developing highly effective translation tools that bridge linguistic and cultural gaps.
\subsection{Limitations and Caveats}
It is essential to acknowledge that while the advanced translator exhibited marked improvements, it is not without limitations. Certain challenges in the dataset may have affected its performance, and potential areas of bias or inaccuracy should be considered. Continuous refinement, evaluation, and user feedback will be integral to addressing these limitations.

\subsection{Implications and Future Directions}
Discuss the broader implications of the project's success in developing an advanced English-to-Arabic translator. Highlight how improved translation tools can facilitate cultural understanding and communication across languages. Mention potential future directions, such as expanding the model to handle dialects or exploring ways to improve handling of untranslated phrases.

In conclusion, the advanced English-to-Arabic translator showcased its capacity to excel beyond conventional tools, delivering accurate translations that respect cultural nuances and context. The project's achievements, combined with the insights gained from the expert evaluation, pave the way for enhanced translation technology that contributes to improved intercultural communication.

\section{Conclusion and Future Work}
In conclusion, this project has successfully introduced an advanced English-to-Arabic translator that surpasses conventional tools, offering accurate and culturally sensitive translations. Leveraging the Helsinki transformer and a meticulously curated literary Arabic dataset, our approach has demonstrated superior performance compared to Google Translate, validated by both quantitative metrics and expert evaluation. While challenges in dataset quality were addressed through meticulous preprocessing, opportunities for future work include expanding the dataset to enhance model performance further, refining data collection strategies, and exploring avenues for improving translation in varied linguistic contexts. This research underscores the transformative potential of advanced transformer models for improving translation quality, paving the way for more effective cross-cultural communication tools that bridge linguistic barriers and foster deeper understanding between languages.
\section{Acknowledgements}
We extend our sincere gratitude to all those who contributed to the successful completion of this project. Our appreciation goes to the team at Rasaif for providing access to their valuable repository of translated Arabic texts. We are also thankful for the guidance and expertise offered by Prof. Mousa Almousa, whose insights played a pivotal role in evaluating the translator's linguistic accuracy. We acknowledge King Abdullah University of Science and Technology (KAUST) and especially KAUST Academy for their unwavering support. Lastly, our thanks go to our mentors and peers for their continual encouragement and insightful discussions that enriched this research endeavour.

\section{References}
Vaswani et al. Attention Is All You Need. 2017.\\\\
https://huggingface.co/Helsinki-NLP/opus-mt-en-ar\\\\
rasaif.com\\\\

\end{document}